\documentclass[review]{elsarticle}

\usepackage{lineno,hyperref}
\usepackage{times}
\usepackage{latexsym}

\usepackage{graphicx}
\usepackage{amsmath}
\usepackage{booktabs}
\usepackage{algorithm}
\usepackage{algorithmic}
\usepackage{tabularx}
\usepackage{todonotes}

\usepackage{amssymb}
\usepackage{amsthm}
\usepackage{mathtools}

\usepackage{subfigure}
\usepackage{url}
\usepackage{enumerate}
\usepackage{enumitem}

\usepackage{color}

\usepackage{url}
\usepackage{scrextend}

\usepackage{cleveref}

\usepackage[symbol]{footmisc}

\modulolinenumbers[5]

\journal{\_\_ (under review)}

\begin{document}

\begin{frontmatter}

\title{X-ToM: Explaining with Theory-of-Mind for Gaining Justified Human Trust}


\author{Arjun R. Akula$^{1}$\footnote[1]{\label{corauth} Corresponding Author} \footnote[2]{\label{email}Email Addresses: \href{mailto:aakual@ucla.edu}{aakual@ucla.edu} (A.Akula), \href{mailto:liucs.msu@gmail.com}{liucs.msu@gmail.com} (C. Liu),
\href{mailto:sadiyasa@cse.msu.edu}{sadiyasa@cse.msu.edu} (S. Sadiya), 
\href{mailto:hongjing@ucla.edu}{hongjing@ucla.edu} (H. Lu), \href{mailto:sinisa@oregonstate.edu}{sinisa@oregonstate.edu} (S. Todorovic), \href{mailto:jchai@cse.msu.edu}{jchai@cse.msu.edu} (J.Y. Chai), \href{mailto:sczhu@stat.ucla.edu}{sczhu@stat.ucla.edu} (S.C. Zhu)}, 
      Changsong Liu$^{1}$\footref{corauth}\footref{email},
      Sari Saba-Sadiya$^{3}$\footref{corauth}\footref{email},
      Hongjing Lu$^{1}$\footref{corauth}\footref{email},
      \\Sinisa Todorovic$^{2}$\footref{corauth}\footref{email}, Joyce Y. Chai$^{3}$\footref{corauth}\footref{email}, 
      Song-Chun Zhu$^{1}$\footref{corauth}\footref{email} 
     }

\address{$^{1}$Center for Vision, Cognition, Learning, and Autonomy,\\ University of California, Los Angeles, CA 90025 USA }
\address{$^{2}$Oregon State University, Corvallis, OR 97331 USA}
\address{$^{3}$Michigan State University, East Lansing, MI 48824 USA}








\begin{abstract}
We present a new explainable AI (XAI) framework aimed at increasing justified human trust and reliance in the AI machine through explanations. We pose explanation as an iterative communication process, i.e. dialog, between the machine and human user. More concretely, the machine generates sequence of explanations in a dialog which takes into account three important aspects at each dialog turn: (a) human's intention (or curiosity); (b) human's understanding of the machine; and (c) machine's understanding of the human user. To do this, we use Theory of Mind (ToM) which helps us in explicitly modeling human's intention, machine's mind as inferred by the human as well as human's mind as inferred by the machine. In other words, these explicit mental representations in ToM are incorporated to learn an optimal explanation policy that takes into account human's perception and beliefs. Furthermore, we also show that ToM facilitates in quantitatively measuring justified human trust in the machine by comparing all the three mental representations.

We applied our framework to three visual recognition tasks, namely, image classification, action recognition, and human body pose estimation. We argue that our ToM based explanations are practical and more natural for both expert and non-expert users to understand the internal workings of complex machine learning models. To the best of our knowledge, this is the first work to derive explanations using ToM. Extensive human study experiments verify our hypotheses, showing that the proposed explanations significantly outperform the state-of-the-art XAI methods in terms of all the standard quantitative and qualitative XAI evaluation metrics including human trust, reliance, and explanation satisfaction.

\end{abstract}

\begin{keyword}
Explainable Artificial Intelligence, Theory of Mind, Interpretability.
\end{keyword}

\end{frontmatter}


\section{Introduction}
\subsection{Motivation and Objective}
From low risk environments such as movie recommendation systems and chatbots to high risk environments such as self-driving cars, drones, military applications and medical-diagnosis and treatment, Artificial Intelligence (AI) systems are becoming increasingly ubiquitous~\cite{chancey2015role,gulshan2016development,lyons2017certifiable,DBLP:journals/corr/MnihKSGAWR13}. AI is finding its way into a wide array of applications in education, finance, healthcare, telecommunication, and law enforcement. In particular, AI systems built using black box machine learning (ML) models -- such as deep neural networks and large ensembles~\cite{lipton2016mythos,ribeiro2016should,miller2017explanation,yang2018,sundararajan2017axiomatic,ramprasaath2016grad,zeiler2014visualizing,smilkov2017smoothgrad,kim2014bayesian} -- perform remarkably well on a broad range of tasks and are gaining widespread adoption. However understanding the behavior of these systems remains a significant
challenge as they cannot explain why they reached a specific recommendation or a decision. This is especially problematic in high risk environments such as banking, healthcare, and insurance, where AI decisions can have significant consequences. Therefore, much hope rests on explanation methods as tools to understand the decisions made by these AI systems. 

\begin{figure}[h]
\centering
  \includegraphics[width=0.93\linewidth,height=0.5\linewidth]{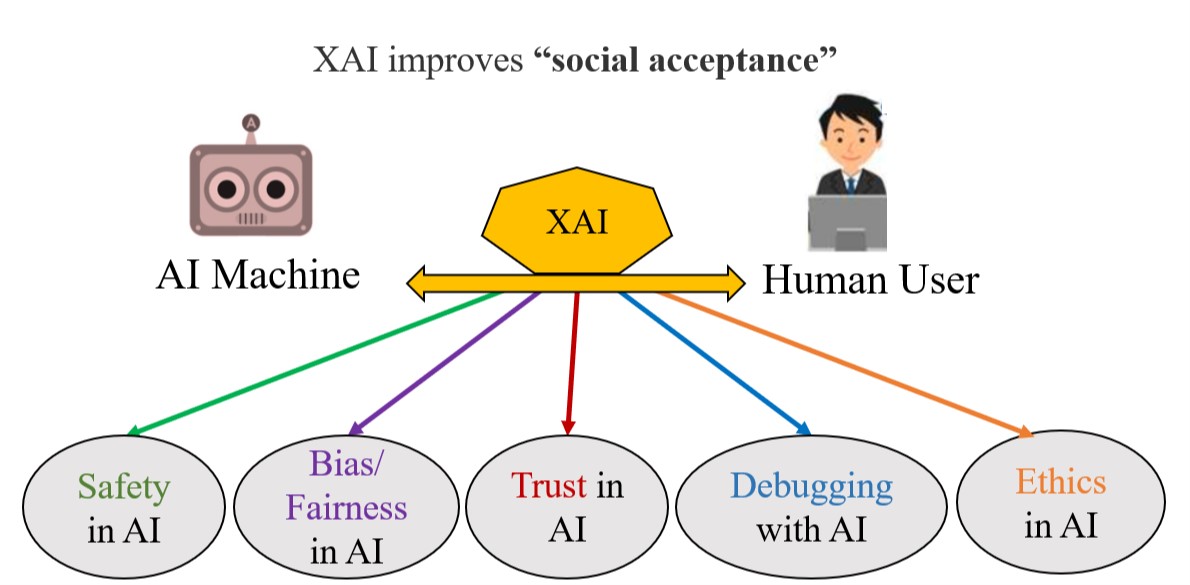}
  \caption{An AI machine that explains its predictions to human users will find more social acceptance. Therefore, XAI models are the key in addressing the issues such as Safety in AI, Bias/Fairness in AI, Trust in AI, Model Debugging, and Ethics in AI. }~\label{fig:fig1}
  \vspace{-24pt}
\end{figure}

Explainable AI (XAI) models, through explanations, make the underlying inference mechanism of AI systems transparent and interpretable to expert users (system developers) and non-expert users (end-users)~\cite{lipton2016mythos,ribeiro2016should,miller2017explanation,hoffman17explanation}. Explanations play a key role in integrating AI machines into our daily lives, i.e. XAI is essential to increase social acceptance of AI machines. As the decision making is being shifted from humans to machines, \textbf{transparency} and \textbf{interpretability} achieved with reliable explanations is central to solving AI problems such as Safety (e.g. \textit{how to operate self-driving cars safely}), Bias \& Fairness (e.g. \textit{how to detect and mitigate bias in ML models}), Justified Human Trust in ML models (e.g. \textit{how to trust the output of these AI systems to inform our decisions}), Model Debugging (e.g. \textit{how to improve my model by identifying points of
model failure}), and Ethics (e.g. \textit{how to ensure that ML models reflect our values}) (Figure~\ref{fig:fig1}).

In this work, we focus mainly on measuring and increasing \textbf{Justified Positive Trust} (JPT) and \textbf{Justified Negative Trust} (JNT)~\cite{hoffman2018metrics} in AI systems. We measure JPT and JNT by evaluating the human’s understanding
 of the machine’s (M) decision-making process. For example, let us consider an image classification task. Suppose if the machine M predicts images in the set $C$ correctly and makes incorrect decisions on the images in the set $W$. Intuitively, JPT will be computed as the percentage of images in $C$ that the human subject felt $M$ would correctly predict. Similarly, JNT (also called as mistrust), will be computed as the percentage of images
in $W$ that the human subject felt $M$ would fail to predict correctly. Note that this definition of justified positive and negative trust is domain generic and can be applied to any task. For example, in an AI-driven clinical world, our definitions of JPT and JNT can effectively measure how much doctors and patients understand the AI systems that assist in clinical decisions. 





\subsection{Introducing X-ToM: Explaining with Theory-of-Mind for Increasing JPT and JNT}
Our work is motivated by the following three key observations:
\begin{enumerate}
    \item \textbf{Attention is not a Good Explanation:} Previous studies have shown that trust is closely and positively correlated to the level of how much human users understand the AI  system --- {\em understandability} --- and how accurately they can predict the system's performance on a given task --- {\em predictability}~\cite{hoffman17explanation,lipton2016mythos,hoffman2018metrics,miller2017explanation}.  Therefore there has been a growing interest in developing explainable AI systems (XAI) aimed at increasing understandability and predictability by providing explanations about the system's predictions to human users~\cite{lipton2016mythos,ribeiro2016should,miller2017explanation,yang2018}. Current works on XAI generate explanations about their performance in terms of, e.g., feature visualization and attention maps~\cite{sundararajan2017axiomatic,ramprasaath2016grad,zeiler2014visualizing,smilkov2017smoothgrad,kim2014bayesian,zhang2018interpretable}. However, solely generating explanations, regardless of their type (visualization or attention maps) and utility, {\em is not sufficient} for increasing understandability and predictability~\cite{DBLP:journals/corr/abs-1902-10186}. We verify this in our experiments (see Section 4). 

\item \textbf{Explanation is an Interactive Communication Process:} We believe that an effective explanation cannot be one shot and involves iterative process of communication between the human and the machine. The context of such interaction plays an important role in determining the utility of the follow-up explanations~\cite{clark1989contributing}. As humans can easily be overwhelmed with too many or too detailed explanations, interactive communication process helps in understanding the user and identify user-specific content for explanation. Moreover, cognitive studies~\cite{miller2017explanation} have shown an explanation can only be optimal if it is generated by taking user's perception and belief into account.  

\begin{figure}[h]
\centering
  \includegraphics[width=0.75\linewidth]{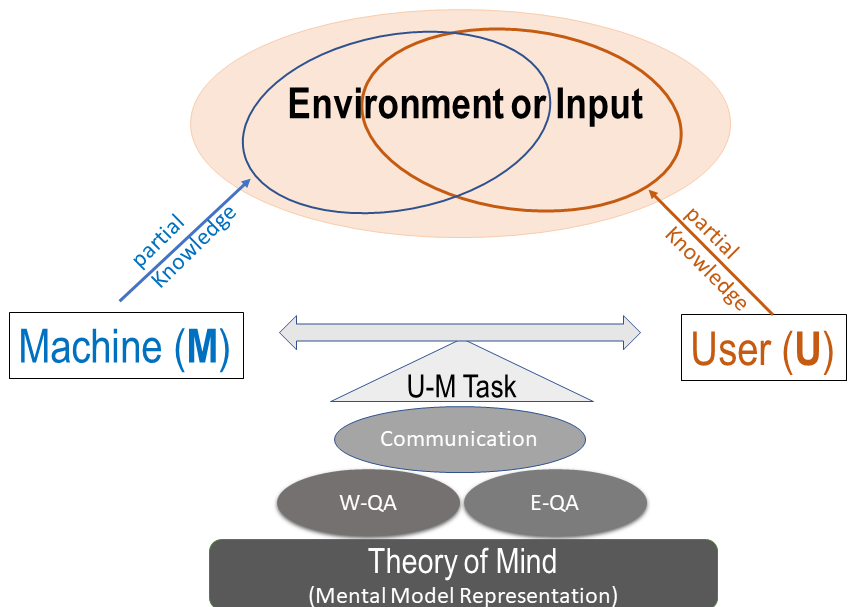}
  \caption{\textbf{XAI as Collaborative Task Solving}: Our interactive and collaborative XAI framework based on the Theory of Mind. The interaction is conducted through a dialog where the user poses questions about facts in the environment (W-QA) and explanation seeking questions (E-QA).}~\label{fig:2agents}
\end{figure}

\item \textbf{Defining a Collaborative Task for the Communication Process:} In our experiments, we found that it is difficult to evaluate the effectiveness of explanations without constraining the communication process. In our framework, we constrain the communication by explicitly defining a collaborative task for the human user to solve through the explanations. Based on how many tasks that are successfully solved by the user (and the number of explanations in the dialog), we measure the effectiveness of the explanations. 
\end{enumerate}

\subsubsection{X-ToM Framework}
Based on the above three key observations, we introduce an interactive explanation framework, \textbf{X-ToM}. In our framework, the machine generates sequence of explanations in a dialog which takes into account three important aspects at each dialog turn: (a) human's intention (or curiosity); (b) human's understanding of the machine; and (c) machine's understanding of the human user. To do this, we use Theory of Mind (ToM) which helps us in explicitly modeling human's intention, machine's mind as inferred by the human as well as human's mind as inferred by the machine. The ability to reason about other's perception and beliefs, in addition to one's own perception and beliefs, is often referred to as the Theory-of-Mind~\cite{devin2016implemented,goldman2012oxford,premack1978does}.

More specifically, in X-ToM, the machine and the user are positioned to solve a collaborative task, but the machine's mind ($M$) and the human user's mind ($U$) only have a partial knowledge of the environment (see Figure~ref\ref{fig:2agents}). Hence, the machine and user need to communicate with each other, using their partial knowledge, otherwise they would not be able to optimally solve the collaborative task. The communication consists of two different types of question-answer (QA) exchanges --- namely, a) Factoid question-answers about the environment (W-QA), where the user asks ``WH''-questions that begin with  \texttt{what}, \texttt{which}, \texttt{where}, and \texttt{how}; and b) Explanation seeking question-answers (E-QA), where the user asks questions that begin with \texttt{why} about the machine's inference. At each turn in the collaborative dialog, our X-ToM updates a model of human perception and beliefs, and uses this model for optimizing explanations in the next turn.


We argue that our interactive explanation framework based on ToM is practical and more natural for both expert and non-expert users to understand the internal workings of complex machine learning models. Furthermore,  we also show that ToM facilitates in quantitatively measuring justified human trust in the machine by comparing all the three mental representations. To the best of our knowledge, this is the first work to derive explanations using ToM.

We applied our framework to three visual recognition tasks, namely, image classification, action recognition, and human body pose estimation. Using Amazon Mechanical Turk, we have collected explanation dialogs by interacting with turkers through X-ToM framework. From there, X-ToM learned an optimal explanation policy that takes into account user perception and beliefs. Through our extensive human studies, we show that X-ToM allows the user to achieve a high success rate in visual recognition on blurred images, and does so very efficiently in a few dialog exchanges. We also found that the most popularly used attribution based explanations (viz. saliency maps) are not effective to improve human trust in AI system, whereas our Theory-of-Mind inspired approach significantly improves human trust in AI by providing effective explanations.

\subsection{Related Work}
Generating explanations or justifications of predictions or decisions made by an AI system has been widely explored in AI. Most prior work has focused on generating explanations using feature visualization and attribution.

\textbf{Feature visualization} techniques typically identify  qualitative interpretations of  features used for making predictions or decisions. 
Recently, there has been an increased interest in developing feature visualizations for deep learning models, especially for  Convolutional Neural Nets (CNNs) in computer vision applications, and  Recurrent Neural Nets (RNNs) in NLP applications.
For example, gradient ascent optimization is used in the image space to visualize the hidden feature layers of unsupervised deep architectures~\cite{erhan2009visualizing}. Also, convolutional layers are visualized by reconstructing the input of each layer from its output~\cite{zeiler2014visualizing}. Recent visual explanation models seek to jointly classify the image and explain why the predicted class label is appropriate for the image~\cite{hendricks2016generating}. Other related work includes a visualization-based explanation framework for Naive Bayes classifiers~\cite{greiner2003explaining}, an interpretable  character-level language models for analyzing the predictions in RNNs~\cite{karpathy2015visualizing}, and an interactive visualization for facilitating analysis of RNN hidden states \cite{strobelt2016visual}.

\textbf{Attribution} is a set of techniques that highlight pixels of the input image (saliency maps) that most caused the output classification. Gradient-based visualization methods~\cite{zhou2016learning,selvaraju2016grad} have been proposed to extract image regions responsible for the network output. The LIME method proposed by~\cite{ribeiro2016should} explains  predictions of any classifier by approximating it locally with an interpretable model.  Influence measures~\cite{datta2015influence} have been used to identify the importance of features in affecting the classification outcome for individual data points.

More recently, apart from feature visualization and attribution techniques, other important lines of research in explainable AI explore dimensionality reduction techniques~\cite{brinton17framework,maaten2008visualizing} and focus on building models which are intrinsically interpretable~\cite{zhang2017interpretable,stone2017teaching}. There are few recent works in the XAI literature that go beyond the pixel-level explanations. For example, the TCAV technique proposed by \cite{kim2018interpretability} aims to generate explanations based on high-level user defined concepts. Contrastive explanations are proposed by \cite{dhurandhar2018explanations} to identify minimal and sufficient features to justify the classification result. \cite{Goyal2019counterfactual} proposed counterfactual visual explanations that identify how the input could change such that the underlying vision system would make a different decision. More recently, few methods have been developed for building models which are intrinsically interpretable~\cite{zhang2017interpretable}. In addition, there are several works~\cite{miller2017explanation,hilton1990conversational,lombrozo2006structure} on the goodness measures of explanation which aim to understand the underlying characteristics of explanations.

\subsection{Contributions}
The contributions of this work are threefold: (i) a new interactive XAI framework based on the Theory-of-Mind; (ii) a new collaborative task-solving game in the domain of visual recognition for learning collaborative explanation strategies; and (iii) a new objective measure of trust and quantitative evaluation of how humans gain increased trust in a given vision system.

\section{X-ToM Framework} \label{framework}
Our X-ToM consists of three main components: 
\begin{itemize}
    \item A {\bf Performer} that generates image interpretations (i.e., machine's mind represented as $pg^M$) using a set of computer vision algorithms;
    \noindent
    \item An {\bf Explainer} that generates maximum utility explanations in a dialog with the user by accounting for $pg^{M}$ and  $pg^{UinM}$ using reinforcement learning;
    \noindent
    \item An {\bf Evaluator} that quantitatively evaluates the effect of explanations on the human's understanding of the machine's behaviors (i.e., $pg^{MinU}$) and measures human trust by comparing $pg^{MinU}$ and $pg^{M}$. 
\end{itemize}

\subsection{X-ToM Game} \label{xtom-game}
\begin{figure}[t]
\centering
  \includegraphics[width=0.9\linewidth,height=6.5cm]{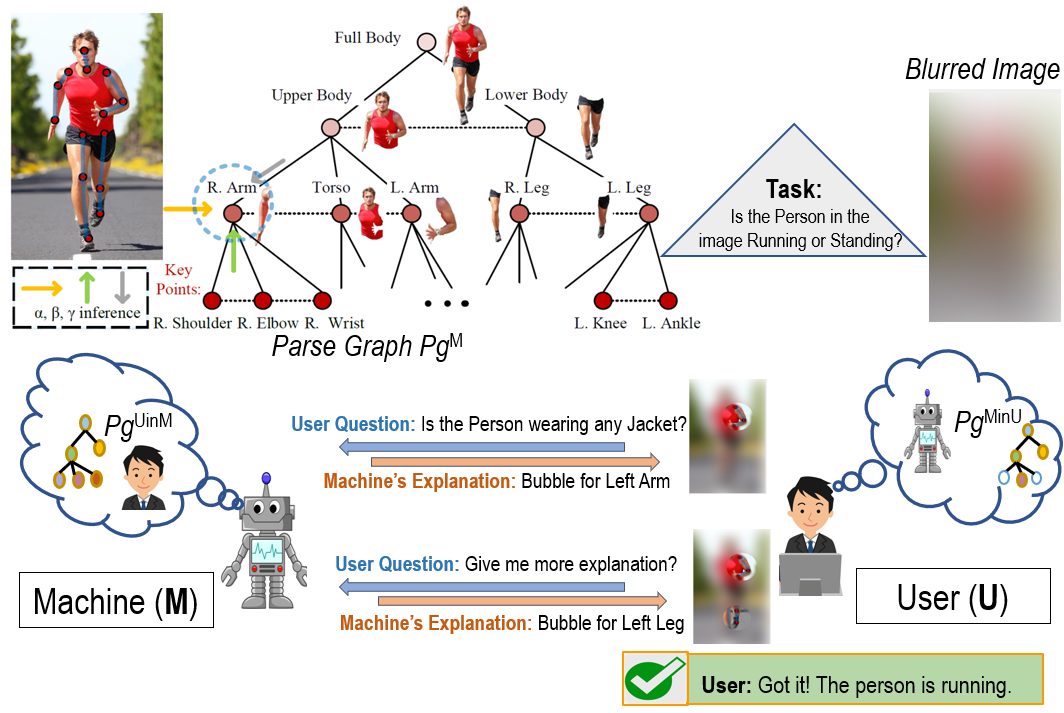}
  \caption{An example of the first phase of an X-ToM game aimed at estimating $pg^{UinM}$: The user is shown a blurred image and given a task to recognize if the person in the image is running or walking. X-ToM has access to the original (unblurred) image and $pg^M$. The user then asks questions regarding objects and parts in the image. Using the detections in $pg^M$, X-ToM provides visual explanations as ``bubbles'' that reveal the corresponding image parts in the blurred image. The generated explanations are used to update $pg^{UinM}$.}~\label{fig:xtom}

\end{figure}

An X-ToM game consists of two phases. 
The first phase is the collaborative task phase. 
The user is shown a blurred image and given a task to recognize what the image shows. X-ToM has access to the original (unblurred) image and the machine's (i.e. \textbf{Performer's}) inference result $pg^M$ (see Section~\ref{xtom-performer}). The user is allowed to ask questions regarding objects and parts in the image that the user finds relevant for his/her own recognition task. Using the detected objects and parts in $pg^M$, X-ToM \textbf{Explainer} provides visual explanations to the user, as shown in Figure~\ref{fig:xtom}. This process allows the machine to infer what the user sees and iteratively update $pg^{UinM}$, and thus select an optimal explanation at every turn of the game (see Section~\ref{xtom-explainer}). Optimal explanations generated by the \textbf{Explainer} are the key to maximize the human trust in the machine. 
The second phase is specifically designed for evaluating whether the explanation provided in the first phase helps the user understand the system behaviors. 
The \textbf{Evaluator} shows a set of original (unblurred) images to the user that are similar to (but different from) the ones used in the first phase of the game (i.e., the set of images shows the same class of objects or human activity). The user is then given a task to predict in each image the locations of objects and parts that would be detected by the machine (i.e., in $pg^M$) according to his/her understanding of the machine's behaviors. Based on the human predictions, the \textbf{Evaluator} estimates $pg^{MinU}$ and quantifies human trust in the machine by comparing $pg^{MinU}$ and $pg^{M}$ (see Section~\ref{xtom-evaluator}).

\subsection{X-ToM Performer (for Image Interpretation)} \label{xtom-performer}

In this paper, the visual tasks involve detecting and localizing human body parts, identifying their poses and attributes, and recognizing human actions from a given image. The AOG for this visual domain uses AND nodes to represent decompositions of human body parts into subparts, and OR nodes for alternative decompositions. 
Each node is characterized by attributes that pertain to the corresponding human body part, including the pose and action of the entire body. Also, edges in the AOG capture hierarchical and contextual relationships of the human body parts.

Our AOG-based performer uses three inference processes $\alpha$, $\beta$ and $\gamma$ at each node. Figure~\ref{fig:xtom} shows an example part of the AOG relevant for human body pose estimation~\cite{park2016attribute}. The $\alpha$ process detects nodes (i.e., human body parts) of the AOG directly based on image features, without taking advantage of the surrounding context. The $\beta$ process infers nodes of the AOG by binding the previously detected children nodes in a bottom-up fashion, where the children nodes have been detected by the $\alpha$ process (e.g., detecting human's upper body from the detected right arm, torso, and left arm).  Note that the $\beta$ process is robust to partial object occlusions as it can infer an object from its detected parts. 
The $\gamma$ process infers a node of the AOG top-down from its previously detected parent nodes, where the parents have been detected by the $\alpha$ process (e.g., detecting human's right leg from the detected outline of the lower body). The parent node passes contextual information so that the performer can detect the presence of an object or part from its surround. Note that the $\gamma$ process is robust to variations in scale at which objects appear in images.


\subsection{X-ToM Explainer (for Explanation Generation)} \label{xtom-explainer}

\begin{figure}[!ht]
\centering
  \includegraphics[width=\linewidth,height=11cm]{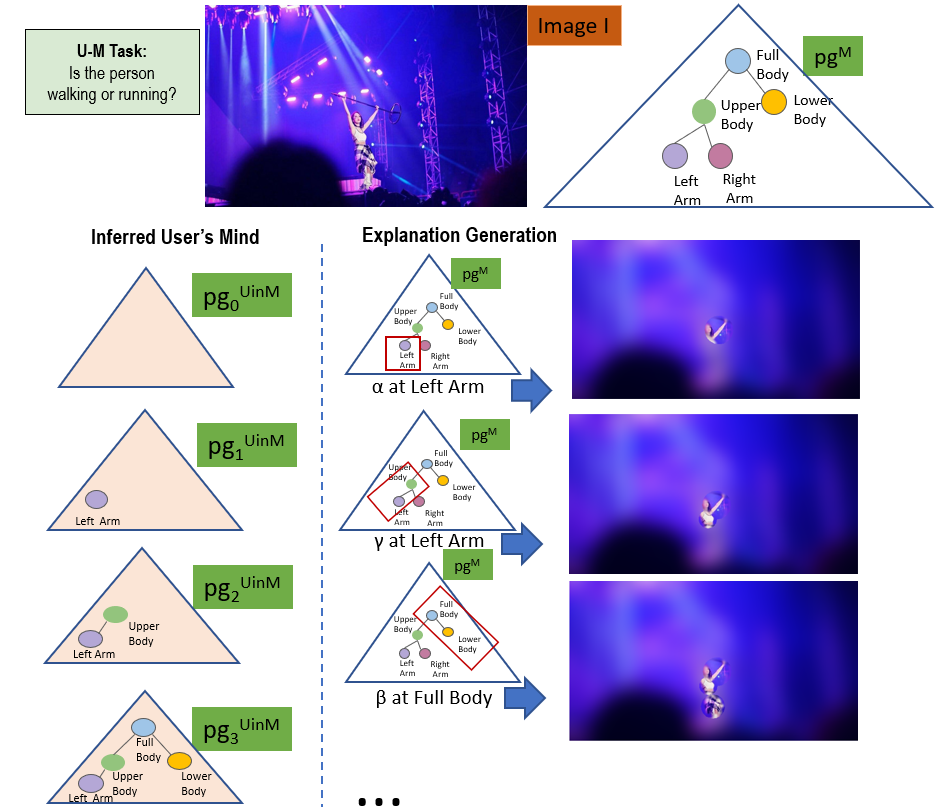}
  \caption{Illustration of the first phase in X-ToM game. The human is asked to solve the task ``Is the person in the image walking or running?". The human may ask questions related to body parts and body poses. The machine reveals a bubble (of various sizes and scales) for each of those questions. The figure shows examples of explanations generated using $\alpha$, $\beta$ and $\gamma$ processes and the updated inferred user's mind after each explanation.}~\label{fig:xtom-eg}
\end{figure}

The explainer, in the first phase of the game, makes the underlying $\alpha$, $\beta$, and $\gamma$ inference process of the performer more transparent to the human through a collaborative dialog. At one end, the explainer is provided access to an image and the performer's inference result $pg^{M}$ on that image. At the other end, the human is presented a blurred version of the same image, and asked to recognize a body part, or pose, or human action depicted (e.g., whether the person is running or walking). To solve the task, the human may ask the explainer various ``what'', ``where'' and ``how'' questions (e.g.,  ``Where is the left arm in the image''). We make the assumption that the human will always ask questions that are related to the task at hand so as to solve it efficiently. The explainer answers these questions using $pg^{M}$ and justifies the answers by showing the corresponding visual explanations in the image (as illustrated in Figure~\ref{fig:xtom-eg}). \\ \\
As visual explanations, we use ``bubbles" \cite{gosselin2001bubbles}, where each bubble reveals a circular part of the blurred image to the human. The bubbles coincide with relevant image parts for answering the question from the human, as inferred by the performer in $pg^{M}$. For example, a bubble may unblur the person's left leg in the blurred image, since that image part has been estimated in $pg^{M}$ as relevant for recognizing the human action ``running'' occurring in the image.

Following the ``principle of least collaborative effort"~\cite{clark1986referring} and the aforementioned findings~\cite{miller2017explanation} that explanations should {\em not}  overwhelm the human, our X-ToM explainer utilizes $pg^{M}$ and  $pg^{UinM}$ (i.e., the contextual and hierarchical relationships explicitly modeled in the AOG) for controlling the depth and breadth of explanations. To enable this control,  each bubble is characterized by a number of parameters, including the amount of image reveal (i.e., the unblurring level), size, and location in the image, to name a few. We use reinforcement learning to train the explainer to optimize these parameters and thus provide optimal visual explanations.

\subsection{X-ToM Evaluator (for Trust Estimation)} \label{xtom-evaluator}


The second phase of the X-ToM game serves to assess the effect of the explainer on the human's understanding of the performer. This assessment is conducted by the evaluator. The human is presented with a set of (unblurred) images that are different from those used in the first phase. For every image, the evaluator asks the human to predict the performer's output. The evaluator poses multiple-choice questions and the user clicks on one or more answers (see Appendix 5.2 for more details on evaluator interface and questions). As shown in Figure~\ref{fig:xtom_trust}, we design these questions to capture different aspects of human's understanding of $\alpha$, $\beta$ and $\gamma$ inference processes in the performer. Based on responses from the human, the evaluator estimates $pg^{MinU}$. By comparing $pg^{MinU}$ with the actual machine's mind $pg^{M}$ (generated by the performer), we have defined the following qualitative and quantitative metrics to quantitatively assess human trust~\cite{hoffman17explanation,hoffman2010metrics,hoffman2018metrics,miller2018explanation} in the performer:

\begin{figure}[!t]
\centering
  \includegraphics[width=\linewidth,height=9cm]{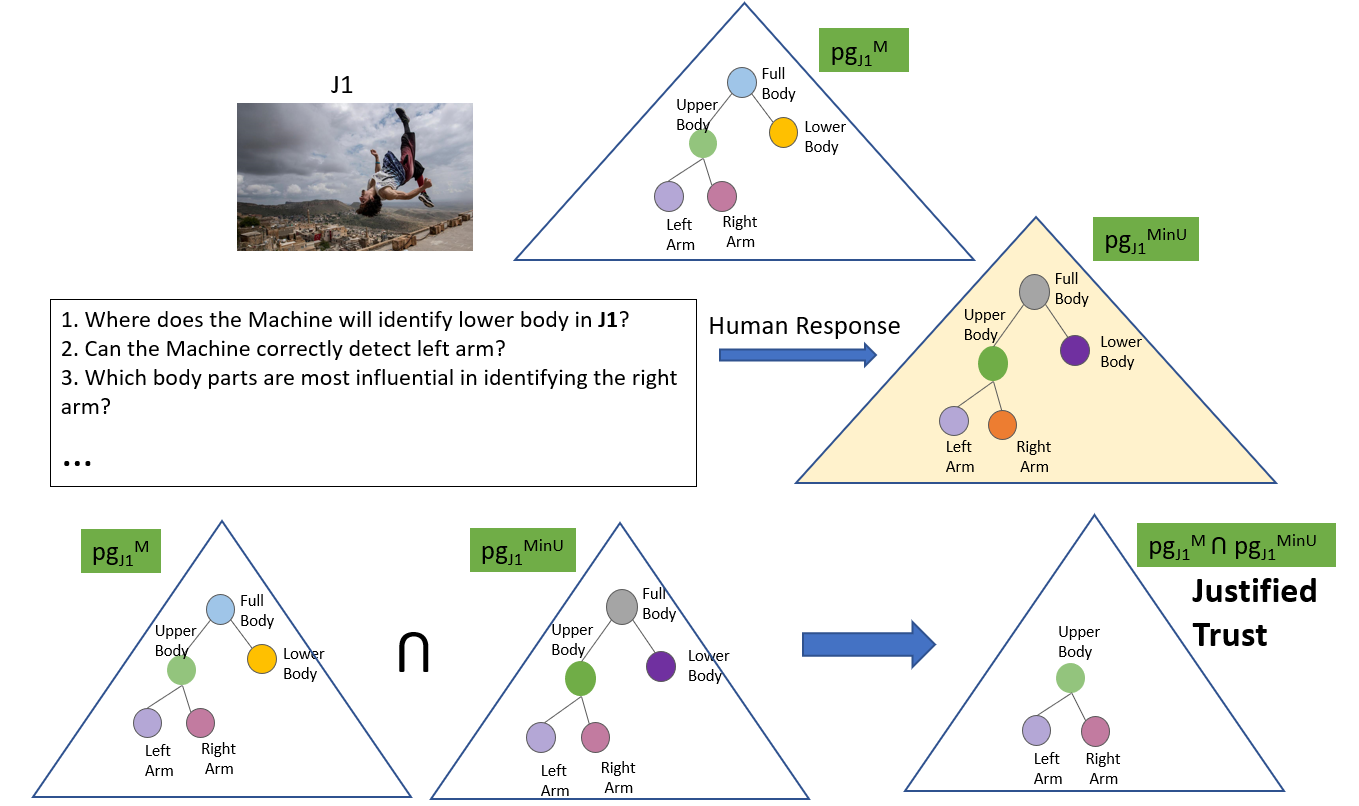}
  \caption{An example of second phase of X-ToM game where we estimate $pg^{MinU}$ and also quantitatively compute justified trust.}~\label{fig:xtom_trust}
\end{figure}

  \noindent
  \textbf{Quantitative Metrics}:\\
 (1) \textit{Justified Positive and Negative Trust:} It is possible for humans to feel positive trust with respect to certain tasks, while feeling negative trust (i.e. mistrust) on some other tasks. The positive and negative trust can be a mixture of justified and unjustified trust~\cite{hoffman17explanation,hoffman2018metrics}. We compute justified positive trust (JPT) and negative trust (JNT) as follows:
 
 \begin{align}
\nonumber
\text{JPT} &= \dfrac{1}{N}\displaystyle\sum_{i}\displaystyle\sum_{z=\alpha,\beta,\gamma}\Delta \text{JPT}(i,z),\\
\nonumber
\Delta \text{JPT}(i,z)  &= \displaystyle\dfrac{\|pg^{MinU}_{i,z,+} \cap pg^{M}_{i,+}\|}{\|pg_{i,+}^{M}\|},\\
\nonumber
\text{JNT} &= \dfrac{1}{N}\displaystyle\sum_{i}\displaystyle\sum_{z=\alpha,\beta,\gamma}\Delta \text{JNT}(i,z),\\
\nonumber
\Delta \text{JNT}(i,z)  &= \displaystyle\dfrac{\|pg^{MinU}_{i,z,-} \cap pg^{M}_{i,-}\|}{\|pg_{i,-}^{M}\|},
\end{align}
\normalsize{}

where $N$ is the total number of games played. $z$ is the type of inference process. $\Delta \text{JPT}(i,z)$, $\Delta \text{JNT}(i,z)$ denote the justified positive and negative trust gained in the $i$-{th} turn of a game on the $z$ inference process respectively. $pg^{MinU}_{i,z,+}$ denotes nodes in $pg^{MinU}_{i}$ for which the user thinks the performer is able to accurately detect in the image using the $z$ inference process. Similarly, $pg^{MinU}_{i,z,-}$ denotes nodes in $pg^{MinU}_{i}$ for which the user thinks the performer would fail to detect in the image using the $z$ inference process. $\|pg\|$ is the size of $pg$. Symbol $\cap$ denote the graph intersection of all nodes and edges from two $pg$'s.

(2) \textit{Reliance:} Reliance (Rc) captures the extent to which a human can accurately predict the performer's inference results without over- or under-estimation. In other words, Reliance is proportional to the sum of JPT and JNT.

\begin{align}
\nonumber
\text{Rc} &= \dfrac{1}{N}\displaystyle\sum_{i}\displaystyle\sum_{z=\alpha,\beta,\gamma}\Delta \text{Rc}(i,z),\\
\nonumber
\Delta\text{Rc}(i,z) &= \displaystyle\dfrac{\|pg^{MinU}_{i,z} \cap pg^{M}_{i,z}\|}{\|pg_{i}^{M}\|}.
\end{align}
\normalsize{}

\noindent
\textbf{Qualitative Metrics}:\\
(3) \textit{Explanation Satisfaction (ES)}. We  measure  users’ feeling of satisfaction at having achieved an understanding of the machine in terms of usefulness, sufficiency, appropriated detail, confidence, accuracy, and consistency. We ask them to rate each of these metrics on a Likert scale of 0 to 9. 

\section{Learning X-ToM Explainer Policy}
Given the following input: image $I$, task $T$ assigned to the human, dialog history $h_i$ of a sequence of generated bubbles, and question from the user $q_i$ selected from a finite set of allowed questions $\mathcal{Q}(T)$ for task $T$, the explainer estimates an optimal explanation  $e_i$ at dialog turn $i$ as

\begin{equation}
\nonumber
\begin{aligned}
 e_i = \operatorname*{arg\,max}_e  \:U\left(e \mid pg^M, pg^{UinM}_i, q_i, h_i, T, I; \theta\right)
\end{aligned}
\end{equation}

where $U$ denotes the utility function parameterized by $\theta$. The set of questions $\mathcal{Q}(T)$ is automatically generated from all concepts (objects, object parts, human activities, object attributes, etc.) that may appear in the image and are also modeled by the Performer. During interaction, the user is prompted to ask a question from this list\footnote{A NLU component can be added to map users' free-form natural language questions to the list of interpretable questions.}. 

\begin{figure*}[t]
\centering
  \includegraphics[width=\linewidth,height=5.5cm]{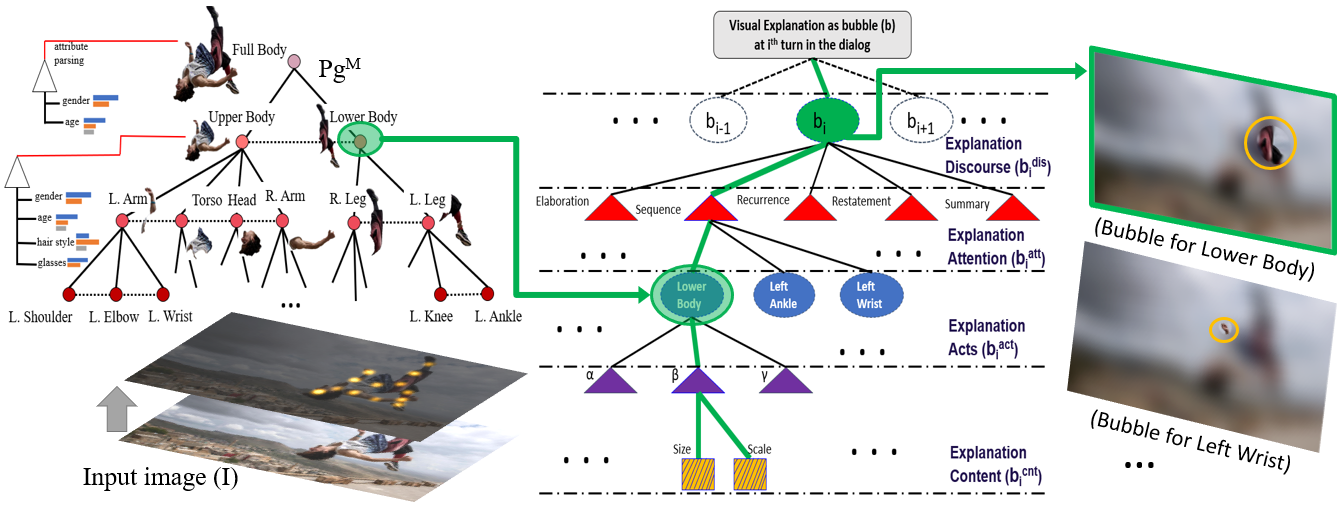}
  \caption{ {\bf Left}: The
Machine interprets the image I as $Pg^M$; {\bf Middle}: Hierarchical representation of the bubble using the four parameters: explanation content ($b^{cnt}$), explanation attention ($b^{act}$), explanation acts ($b^{att}$) and explanation discourse ($b^{att}$); {\bf Right}: The Human receives visual
explanations -- bubbles -- optimized by the X-ToM Explainer.}~\label{fig:figure2}
 
\end{figure*}

As defined earlier, $pg^{UinM}_i$ denotes the current estimate of human's mind, which is an empty graph without nodes and edges at the beginning of the X-ToM game. At every turn in the dialog, the explainer infers and updates $pg^{UinM}_i$ by maximizing its posterior distribution based on $h_i$, $T$ and  $q_i$. Using a Bayesian approach, we define the posterior of $pg^{UinM}_i$ as

\begin{equation}
\nonumber
\begin{aligned}
&p\left(pg^{UinM}_i\mid h_i,q_i,T\right)\propto \\
&p\left(q_i \mid h_i,pg^{UinM}_i,T\right)p\left(h_i \mid pg^{UinM}_i,T\right)p\left(pg^{UinM}_i,T\right)
\end{aligned}
\end{equation}

where $p\left(pg^{UinM}_i, T\right)$ is specified as a uniform prior. The likelihoods $p\left(q_i \mid h_i, , pg^{UinM}_i, T\right)$ and $p\left(h_i \mid  pg^{UinM}_i, T\right)$ are estimated based on the frequency of occurrence of the question $q=q_i$ and the dialog history $h=h_i$ over many X-ToM games played with human users. After updating $pg^{UinM}_i$, the selection of an optimal bubble, i.e., explanation $e_i$, is cast as a sequential decision-making problem and formalized using reinforcement learning (RL). Below we specify the state, actions, reward, and policy of the RL framework. \\
\noindent
\textbf{RL State} ($s_i$). The state of the explainer at dialog turn $i$ consists of $pg^M$, $pg^{UinM}_i$, $q_i$, and  $h_i$.\\
\noindent
\textbf{RL Action} ($a_i$). The action space consists of all  possible bubbles that can be generated from $pg^M$ so that they reveal relevant image parts in the blurred image to the human. Each bubble $b$ is characterized by the following four groups of parameters, as illustrated in Figure~\ref{fig:figure2}:\\ 
\noindent
(a) \textbf{Explanation Content}, $b^{cnt}$, is defined as the amount of visual information contained in the bubble. Our X-ToM uses the Gaussian scale-space \cite{witkin1987scale} for measuring $b^{cnt}$. Specifically, we model ``space'' as a Gaussian with variance $\sigma_1^2$ governing the length of the radius (i.e., spatial size) of the bubble. Also, we model ``scale'' as a Gaussian with variance $\sigma_2^2$ governing the amount of image unblur that the bubble reveals to the user. Given  $\sigma_1^2$ and $\sigma_2^2$, we compute $b^{cnt}_i$ as the differential entropy 

    \begin{equation}
    \nonumber
        b^{cnt}=1+\frac{1}{2}\log(4\pi^2\sigma_1^2\sigma_2^2)
        \label{eq:bcnt}
    \end{equation}

Intuitively, a bubble with large ``space'' (i.e., large size) and large ``scale'' (i.e., high resolution) reveals a lot of information about the image. Conversely, a bubble with small ``space'' and ``scale'' reveals very little evidence. If the explainer always chose bubbles with small ``space'' and ``scale'', it would lead to inefficient dialogue for solving the task. On the other hand, if the explainer always chose bubbles with large space and scale, it would distract the human with unnecessary information and make it difficult for the human to understand the machine's internal representation and inference\footnote{For example, showing a very large bubble for revealing Left-Wrist will also reveal Left-Elbow to the human. This makes it harder for human to understand whether the machine is capable of detecting the exact location of Left-Wrist in the image. In addition, although larger bubbles can potentially minimize the number of turns, they transmit a large amount of information from machine to human. This effect may not be obvious in the current experimental set up, but will be significant in the situation where information to be transmitted is through text. Larger bubbles will correspond to longer textual descriptions.}. Thus, the explainer's goal is to find the bubble with an optimal $b^{cnt}$. In this paper, we discretize ``space'' and ``scale'' of bubbles using  $\sigma_1 \in \{1.15, 3.15, 4.5\}$, and $\sigma_2\in \{1, 9, 15\}$. \\
\noindent
(b) \textbf{Explanation Acts}, $b^{act}$, parametrizes the three types of visual explanations (i.e., bubbles) that can be presented to the human,  corresponding to the three inference processes in our AOG-based performer. Specifically, $b^{act}$ can be: $\alpha$,  $\beta$, or $\gamma$ explanation act. Note that using $\beta$ and $\gamma$ explanation acts (i.e., bottom-up and top-down inference processes of the performer) allows for increasing depth of explanations.\\
\noindent
(c) \textbf{Explanation Attention}, $b^{att}$, indexes a particular human body part from $pg^{M}$ that is the current focus of the dialog with the human. In the paper, the AOG explicitly models human body parts and their subparts, where $pg^{M}$ infers only a subset of those appearing in the image.\\
\noindent
(d) \textbf{Explanation Discourse}, $b^{dis}$, parametrizes discourse relations of the bubbles generated along the dialog with the human. In this paper, we account for the dialog discourse  for enforcing coherence among the explanations. In our experiments, we found the following five discourse relations~\cite{carlson2003building,clark1986referring} to be sufficient and helpful:

 \begin{enumerate}
 \item \textbf{Elaboration}. If  bubble $b_{i+1}$ provides additional details (e.g., by increasing ``scale'' or ``space'') relative to the previous bubbles $h_i=b_{1...i}$, then $b_{i+1}$ relates to the dialog history $h_i$ with the \textit{elaboration} relationship. 
 
\item \textbf{Sequence}. If the explanation attention $b^{att}$ of bubble $b_{i+1}$ is not part of the dialog history $h_i$, then $b_{i+1}$ relates to $h_i$ with the \textit{sequence} relationship.

\item \textbf{Recurrence}. If bubble $b_{i+1}$ already exists in $h_i$, then the discourse relationship between $b_{i+1}$ and $h_i$ is called \textit{recurrence}.

\item \textbf{Restatement}. If the dialog history $h_i$ already contains a bubble with the same explanation attention $b^{att}$ as $b_{i+1}$, then $b_{i+1}$ relates to $h_i$ with the \textit{restatement} relationship.

\item \textbf{Summary} is a special case of the elaboration relationship. If an attention node of $pg^M$ has been already explained in the dialog history $h_i$, and $b_{i+1}$ has the same explanation attention but corresponds to a lower resolution and larger size bubble than the one in $h_i$, then  $b_{i+1}$ relates to  $h_i$ with the \textit{summary} relationship.
\end{enumerate}

\noindent
\textbf{RL Reward} ($r_i$) Our reward function aims to maximize the success rate (ss), user confidence (cf), user satisfaction (sf) and minimize the cost ($C_i$) over the total number of bubbles. We estimate the cost of generating bubbles $b_1$,$b_2$,...,$b_i$ as

\begin{equation}
\nonumber
        C_i = \sum_{j=1}^{i}\frac{1}{b^{cnt}_j}
        \label{eq:cost}
\end{equation}
\noindent

RL Reward ($r_i$) is expressed in terms of a user feedback and cost associated with selecting the bubbles. At each dialog turn $i$, after choosing $b_i$, the explainer collects the following feedback from the user:
\begin{enumerate}
    \item {\em Success} ($\text{ss}_i$): The user is asked to solve the task based on $\{b_i, h_i\}$. The user's success indicates that the machine's dialog with the user had a high utility and the explanations made by the machine make sense and can help the user reach an understanding of the image. Therefore, if the user solves the task correctly, the explainer is rewarded  with $\text{ss}_i$ = 1; otherwise, $\text{ss}_i$ = -1.
    
    \item {\em User confidence} ($\text{cf}_i$): It is possible that user might solve the task by chance without really understanding the task. We therefore additionally ask the user to report their confidence in solving the task on a scale of 1 to 5.
    \item {\em User satisfaction} ($\text{sf}_i$): We ask the user to rate the ordering of bubbles generated in the dialog, and their relevance for solving the task on a scale of 1 to 5.
\end{enumerate}
To compute $r_i$, we also estimate the cost function $C_i$ of generating bubbles $b_1$,$b_2$,...,$b_i$, defined as 
\begin{equation}
        C_i = \sum_{j=1}^{i}\frac{1}{b^{cnt}_j},
        \label{eq:cost}
    \end{equation}
 where  $b^{cnt}$ is computed as follows: \\
 \begin{equation}
        b^{cnt}=1+\frac{1}{2}\log(4\pi^2\sigma_1^2\sigma_2^2).
        \label{eq:bcnt}
    \end{equation}
    
    Intuitively, a large $C_i$ indicates that explanation content of the bubbles revealed is high. 
    
    Our reward function aims to maximize the success rate (ss), user confidence (cf), user satisfaction (sf) and minimize the cost ($C_i$) over the total number of bubbles.  We estimate the cost of generating bubbles $b_1$,$b_2$,...,$b_i$ as
    
\begin{equation}
\begin{multlined}
r_i = \frac{1}{i} \exp(\frac{\text{ss}_i ~ \text{cf}_i ~ \text{sf}_i}{C_i}).
\end{multlined}
\end{equation}

\textbf{RL Policy and Training}. The explainer operates under a stochastic policy, $\pi\left(a_i | s_i;\theta\right)$, which samples optimal bubbles conditioned on the state. This policy is learned by a standard recurrent neural network, called Long-Short Term Memory (LSTM) \cite{hochreiter1997long}. In this paper we use a 2-layer LSTM parameterized by $\theta$. Input to the LSTM is a  feature vector representing the state $s_i$ -- specifically, a binary indicator vector of the AOG nodes and edges present in $pg^M$ and $pg^{UinM}_i$, as well as indices of the question $q_i$ and  bubbles generated in $h_i$. The LSTM's output is the predicted quadruple $(b^{cnt},b^{act},b^{att},b^{dis})$ of $b_{i+1}$. Thus, the goal of the policy learning is to estimate the LSTM parameters  $\theta$. 

We use actor-critic with experience replay for policy optimization~\cite{wang2016sample}. The training objective is to find  $\pi\left(a_i | s_i;\theta\right)$ that maximizes the expected reward $J(\theta)$ over all possible bubble sequences given a starting state. The gradient of  the objective function has the following form:

\begin{equation}
\begin{aligned}
 \nabla_\theta J(\theta) = \mathbb{E}[\nabla_\theta\log\pi_\theta\left(a_i|s_i;\theta\right)A\left(s_i,a_i\right)]
\end{aligned}
\end{equation}

where $A\left(s_i,a_i\right) = Q\left(s_i,a_i\right) - V(s_i)$ is the advantage function~\cite{sutton2000policy}. $Q\left(s_i,a_i\right)$ is the standard Q-function, and $V(s_i)$ is the baseline function aimed at reducing the variance of the estimated gradient. We use the same specifications of $Q\left(s_i,a_i\right)$ and $V(s_i)$ as in \cite{sutton2000policy}. As in \cite{sutton2000policy}, we sample the dialog experiences randomly from the replay pool for training.

\section{Experiments}



We deployed the X-ToM game on the Amazon Mechanical Turk (AMT) and trained the X-ToM Explainer through the interactions with turkers. All the turkers have a bachelor’s degree or higher. We used three visual recognition tasks in our experiments, namely, human body parts identification, pose estimation, and action identification. We used 1000 images randomly selected from Extended Leeds Sports (LSP) dataset~\cite{johnson2010clustered}. Each image is used in all the three tasks. During training, each trial consists of one X-ToM game where a turker solves a given task on a given image. We restrict Turkers from solving a task on an image more than once. In total, about 2400 unique workers contributed in our experiments.

 We performed off-policy updates after every 200 trials, using Adam optimizer~\cite{kingma2014adam} with a learning rate of 0.001 and gradients were clipped at [-5.0, 5.0] to avoid explosion. We used $\epsilon$-greedy policy, which was annealed from 0.6 to 0.0. We stopped the training once the model converged. In our case, the X-ToM policy model converged after interacting with 3500 turkers. All our data and code will be made publicly available.

\noindent
\begin{table}[h]
\begin{center}
\begin{tabular}{ | m{1.7cm} | m{1.5cm}| m{1.5cm} | m{1.8cm}| m{1.5cm} | } 
\hline
Elaboration & Sequence & Recurrence & Restatement & Summary \\ 
\hline
26\% & 48.7\% & 12.6\% & 5.1\% & 7.6\%\\
\hline
\end{tabular}
\caption{Distribution of observed discourse relations in the test trials}
\label{tab:disc}
\end{center}
\end{table}

\noindent
The trained X-ToM Explainer was applied to an additional 500 X-ToM games with AMT turkers for testing.
Table~\ref{tab:disc} shows the percentage of discourse relations among bubbles found in the test interactions. As can be seen, the discourse relation \texttt{sequence} dominates other relations. This indicates that the X-ToM's most common explanation strategy is to prefer a bubble containing new evidence (that was not already shown to the user). Furthermore, the experiment has shown that 55.3\% of the bubbles in the test trials were generated using $\alpha$ explanation act, 23.1\% using $\beta$ explanation act, and 21.6\% using $\gamma$ explanation act. The high percentage of $\beta$ and $\gamma$ explanation acts indicate that contextual evidence is not only helpful for the performer to detect but also for the explainer to explain.

\subsection{AMT Evaluation of X-ToM Explainer}

We conducted an ablation study to quantify the importance of taking the inferred human's mind into account for generating optimal explanations, i.e., the ablated model does not explicitly represent and infer $pg^{UinM}$. Similar to X-ToM, the ablated model was also deployed and trained on AMT. The trained ablated model was again applied to an additional 500 X-ToM games with AMT turkers for testing. Table~\ref{tab:ablation} compares X-ToM Explainer with the ablated model in terms of objective measures such as average success rate (ss), average number of bubbles, average rewards ($r$). X-ToM Explainer significantly outperforms the ablated model ($p < 0.01$) in terms of the overall reward. Although the success rates of both models are similar, the ablated model is found to use a significantly larger number of bubbles, which leads to lower overall reward.

\begin {table}[h]
\begin{center}
\begin{tabular}{ | m{3cm} | m{1.64cm}| m{1cm} | m{1.5cm} | m{0.6cm} |} 
\hline
Model & \#test trials & ss & \#bubbles & r\\ 
\hline
X-ToM & 500 & \textbf{81.3}\% & \textbf{10.5} & \textbf{0.91}\\
\hline
Ablated Model & 500 & 77.1\% & 28 & 0.42\\
\hline
Human Strategy & 100 & 78.9\% & \textbf{6} & 0.62\\
\hline
\end{tabular}
\caption {Comparison of X-ToM with ablated and human baselines}
\label{tab:ablation}
\end{center}
\end{table}

Using an additional 100 X-ToM games on AMT, we further compare the explanations generated by our X-ToM Explainer with the explanations annotated by humans. We asked three graduate students (not the authors), to select the most appropriate bubbles for a given task. Bubbles that have been agreed upon by these three subjects were taken as the best explanations for the given task and image.
In terms of maximizing the reward, we found that X-ToM Explainer performed significantly better than the human strategy of bubble selection ($p<0.01$). However, we found that the average dialog length in the human explanations is 6, while the average dialogue length observed in the X-ToM explanations is 10.5, indicating that there is a possibility to further improve the quality of the X-ToM explanations. We leave this for future exploration.

\subsection{Human Subject Evaluation on Justified Trust}



 Using X-ToM Evaluator, we conduct human subject experiments to assess the effectiveness of the X-ToM Explainer, that is trained on AMT, in increasing human trust through explanations. We recruited 120 human subjects from our institution's Psychology subject pool~\footnote{These experiments were reviewed and approved by our institution's IRB.}. These subjects have no background on computer vision, deep learning and NLP (see Appendix 5.1 for more details). We applied between-subject design and randomly assigned each subject into one of the three groups. One group used X-ToM Explainer, and two groups used the following two baselines respectively: 
 
\begin{itemize}
\item {\bf $\Omega_{\text{QA}}$}: we measure the gains in human trust only by revealing 
the answers for the tasks without providing any explanations to the human.

\item {\bf $\Omega_{\text{Salience}}$}: in addition to the answers, we also provide saliency maps generated using attribution techniques to the human as explanations~\cite{zhou2016learning,selvaraju2016grad}.

\end{itemize} 



Within each group, each subject will first go through an introduction phase where we introduce the tasks to the subjects. Next, they will go through familiarization phase where the subjects become familiar with the machine's underlying inference process (Performer), followed by a testing phase where we apply our trust metrics and assess their trust in the underlying Performer.

\begin{figure}[t]
\centering
  \includegraphics[width=\linewidth,height=6.8cm]{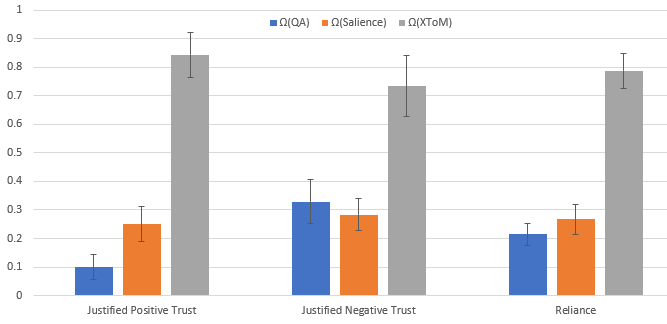}
  \caption{Gain in Justified Positive Trust, Justified Negative Trust and Reliance: X-ToM vs baselines (QA, Saliency Maps). Error bars denote standard errors of the means.}~\label{fig:trust_bars}
\end{figure}

Figure~\ref{fig:trust_bars} compares the justified positive trust (JPT), justified negative trust (JPT), and Reliance (Rc) of X-ToM with the baselines.
As we can see, JPT, JNT and Rc values of X-ToM are significantly higher than $\Omega_{\text{QA}}$ and $\Omega_{\text{Salience}}$ ($p < 0.01$). \textit{Also, it should be noted that attribution techniques ($\Omega_{\text{Salience}}$) did not perform any better than the $\Omega_{\text{QA}}$ baseline where no explanations are provided to the user}. This could be attributed to the fact that, though saliency maps help human subjects in localizing the region in the image based on which the performer made a decision, they do not necessarily reflect the underlying inference mechanism. In contrast, X-ToM Explainer makes the underlying inference processes ($\alpha$, $\beta$, $\gamma$) more explicit and transparent and also provides explanations tailored for individual user's perception and understanding. Therefore X-ToM leads to the significantly higher values of JPT, JNT and Rc. This is one of the key results of our work, given the popularity of attribution techniques as the state-of-the-art explanations.

\begin{figure}[h]
\centering
  \includegraphics[width=\linewidth,height=8cm]{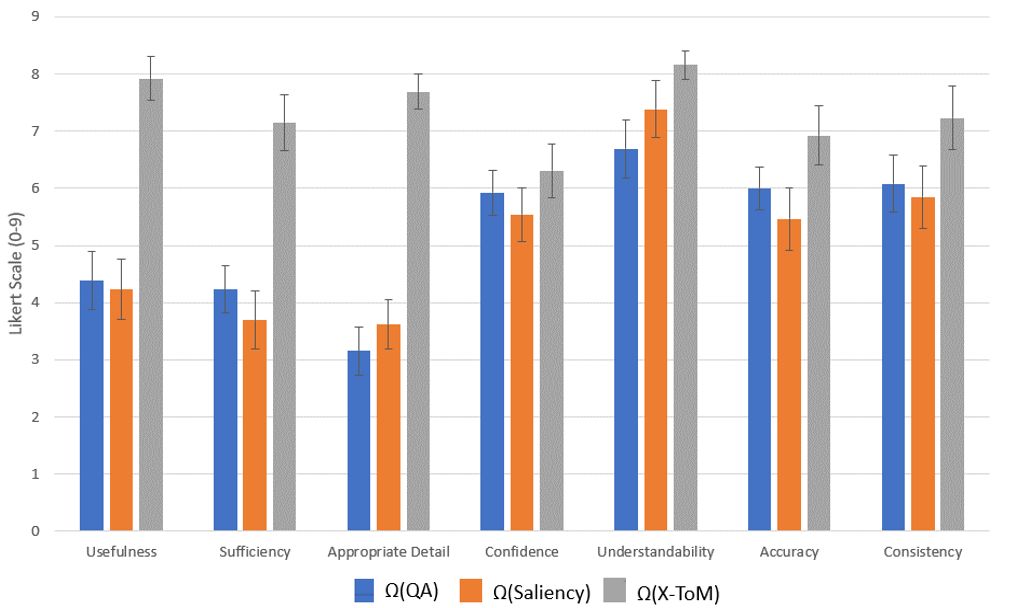}
  \caption{Explanation Satisfaction: X-ToM vs baselines (QA, Saliency Maps). Error bars denote standard errors of the means.}~\label{fig:satisfaction}
\end{figure}

Figure~\ref{fig:satisfaction} shows the average explanation satisfaction rates obtained from each of the three groups. As we can see, subjects in X-ToM experiment group found that explanations were highly useful, sufficient and detailed compared to the baselines ($p < 0.01$). Interestingly, we did not find significant differences across the three groups in terms of other satisfaction measures: confidence, understandability, accuracy and consistency. We leave this observation for future exploration

\begin{figure}[!ht]
\centering
  \includegraphics[width=0.8\linewidth,height=6cm]{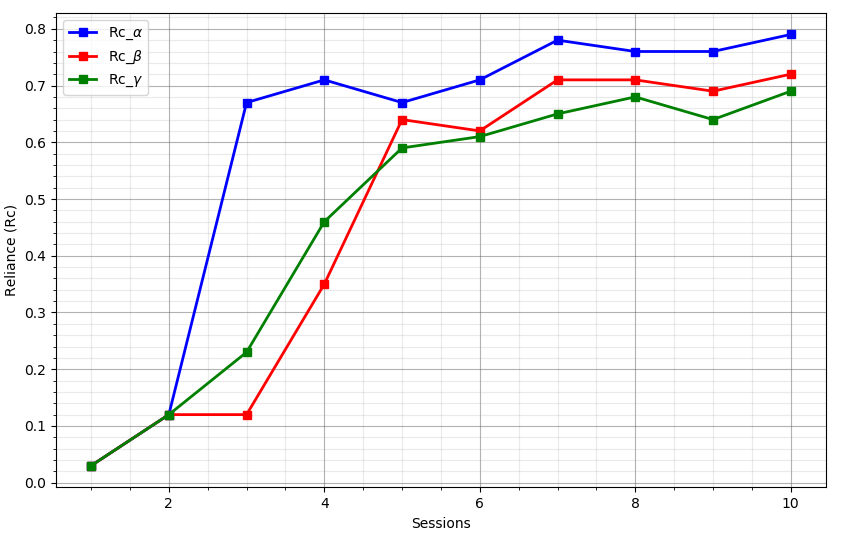}
  \caption{Gain in Reliance over sessions w.r.t $\alpha$, $\beta$ and $\gamma$ processes}~\label{fig:Reliance}
\end{figure}
\begin{figure}[!ht]
\centering
  \includegraphics[width=\linewidth,height=7.1cm]{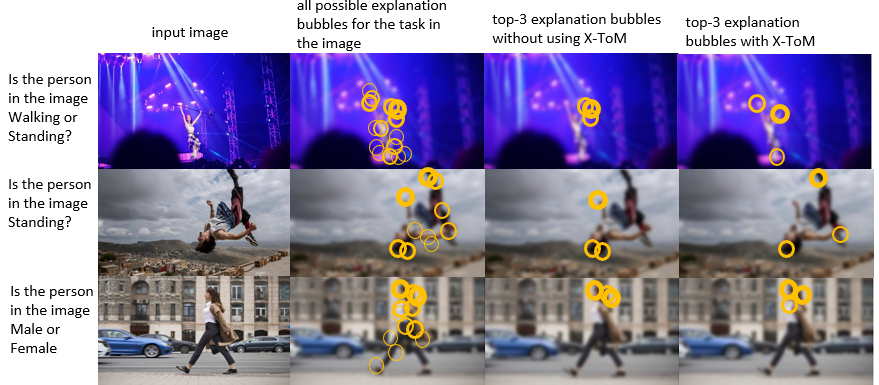}
  \caption{Top-3 best explanations generated with and without using X-ToM.}~\label{fig:case-study}
\end{figure}

\subsection{Gain in Reliance over time}
We hypothesized that human trust and reliance in machine might improve over time. This is because, it can be harder for humans to fully understand the machine's underlying inference process in one single session. Therefore, we conduct an additional experiment with eight human subjects where the subjects' reliance is measured after every session. The results are shown in Figure~\ref{fig:Reliance}. As we expected, subjects' reliance increased over time. Specifically, reliance with respect to $\alpha$ inference process significantly improved only after 2.5 sessions. Reliance with respect to $\beta$ and $\gamma$ inference processes significantly improved after 4.5 sessions. It is clearly evident that, with more sessions, it is possible to further improve human reliance in AI system.

\subsection{Case Study}
Figure~\ref{fig:case-study} shows examples where the top-3 best explanations preferred by X-ToM are compared against the top-3 explanations generated by the attribution techniques. The first column shows the input image for the task. The second column shows all the evidence (i.e., explanations in the form of bubbles, highlighted in yellow color) used in the machine's inference about the task. The thicker the bubble, the higher is its influence, for the machine, in interpreting the image. As we can see, attribution techniques chose the explanations only based on how influential they are for the machine in recognizing the image (third column). In contrast, since X-ToM maximizes the utility of explanations based on both influence values and user's model, explanations selected by the X-ToM (fourth column) are diverse and are more intuitive for humans to understand and solve the task efficiently. For example, for the first image, to aid the human user in solving the task `Is the person in the image walking', X-ToM generates the explanation bubbles based on left arm, right arm and lower body of the person, whereas attribution techniques generate the top-3 bubbles only based on right arm which clearly is not sufficient for the user to successfully solve the task.
\begin{figure}[!ht]
\centering
  \includegraphics[width=0.8\linewidth]{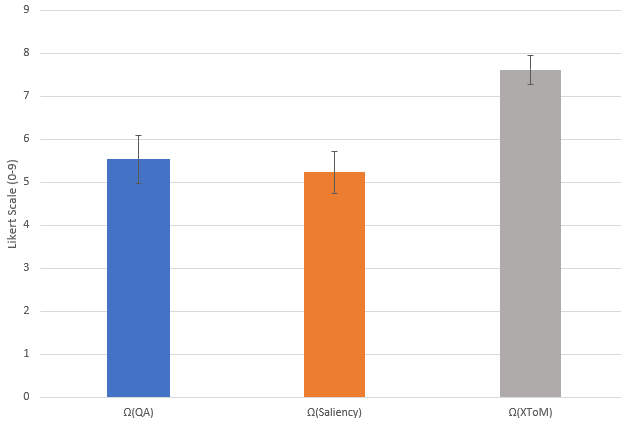}
  \caption{\textbf{Qualitative Reliance}. Error bars denote standard errors of the means.}~\label{fig:subjective}
\end{figure}
\begin{figure}[!ht]
\centering
  \includegraphics[width=0.8\linewidth]{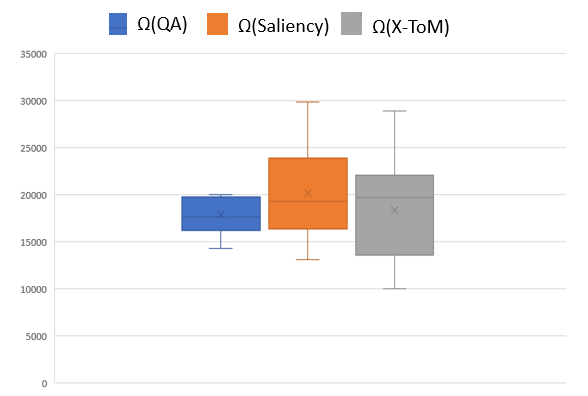}
  \caption{\textbf{Response Times} (in milliseconds per question). Error bars denote standard errors of the means.}~\label{fig:responsetime}
\end{figure}

In addition to the quantitative and qualitative metrics discussed in section 2.5, we also measure the following metrics for comparing our X-ToM framework with the baselines:

\begin{itemize}

\item \textbf{Response Time}: We record the time taken by the human subject in answering evaluator questions. Figure~\ref{fig:responsetime} shows the average response times (in milliseconds per question) for each of the three groups (X-ToM, QA and Saliency Maps). We expected the participants in X-ToM group to take less time to respond compared to the baselines. However, we find no significant difference in the response times across the three groups.

\item \textbf{Subjective Evaluation of Reliance}: We collect subjective Reliance values (on a Likert scale of 0 to 9) from the subjects in the three groups. The results are shown in Figure~\ref{fig:subjective}.  These results are consistent with our quantitative reliance measures. It may be noted that subjects' qualitative reliance in Saliency Maps is lower compared to the QA baseline.
\end{itemize}

%


\section{Conclusions}
This paper presents X-ToM -- a new framework for  Explainable AI (XAI) and human trust evaluation based on the Theory-of-Mind (ToM). X-ToM generates explanations in a dialog by explicitly modeling, learning, and inferring three mental states based on And-Or Graphs -- namely, machine's mind, human's mind as inferred by the machine, and machine's mind as inferred by the human. This allows for a principled formulation of human trust in the machine. For the task of visual recognition, we proposed a novel, collaborative task-solving game that can be used for collecting training data and thus learning the three mental states, as well as a testbed for quantitative evaluation of explainable vision systems. We demonstrated the superiority of X-ToM in gaining human trust relative to baselines.

\section{Acknowledgement}
The work is supported by DARPA XAI N66001-17-2-4029.

\section*{References}


\clearpage
\section{Appendix}

\subsection{Evaluation with Psychology Subject Pool}
\begin{figure}[!ht]
\centering
  \includegraphics[width=\linewidth, height=5.5cm]{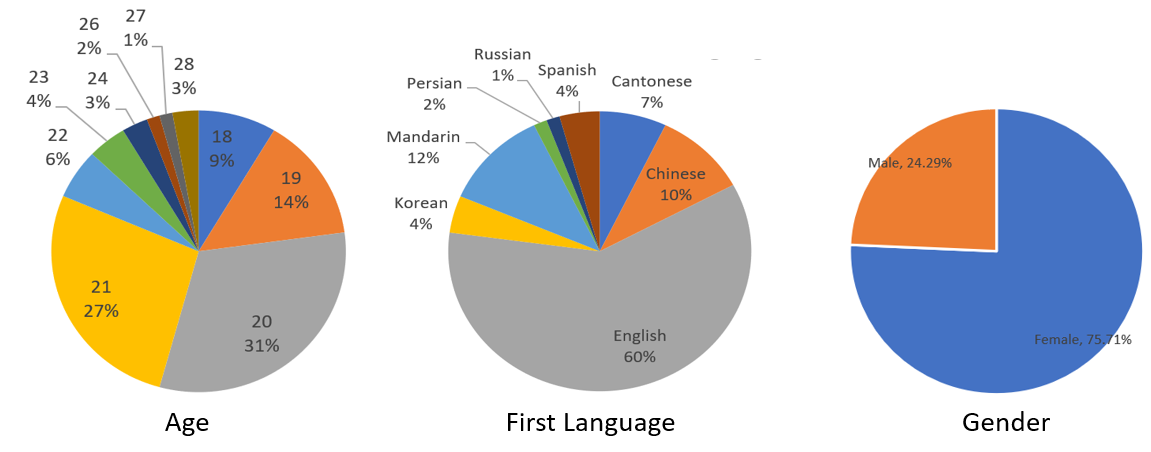}
  \caption{Statistics (based on Age, First Language and Gender) of the 120 human subjects, from Psychology subject pool, participated in our study.}~\label{fig:stats}
\end{figure}

Figure~\ref{fig:stats} shows the statistics (Age, First Language, Gender) of the 120 human subjects, recruited from our institution's Psychology subject pool.

\subsection{X-ToM Evaluator Interface and Questions}
\begin{figure*}[!ht]
\centering
  \includegraphics[width=\linewidth]{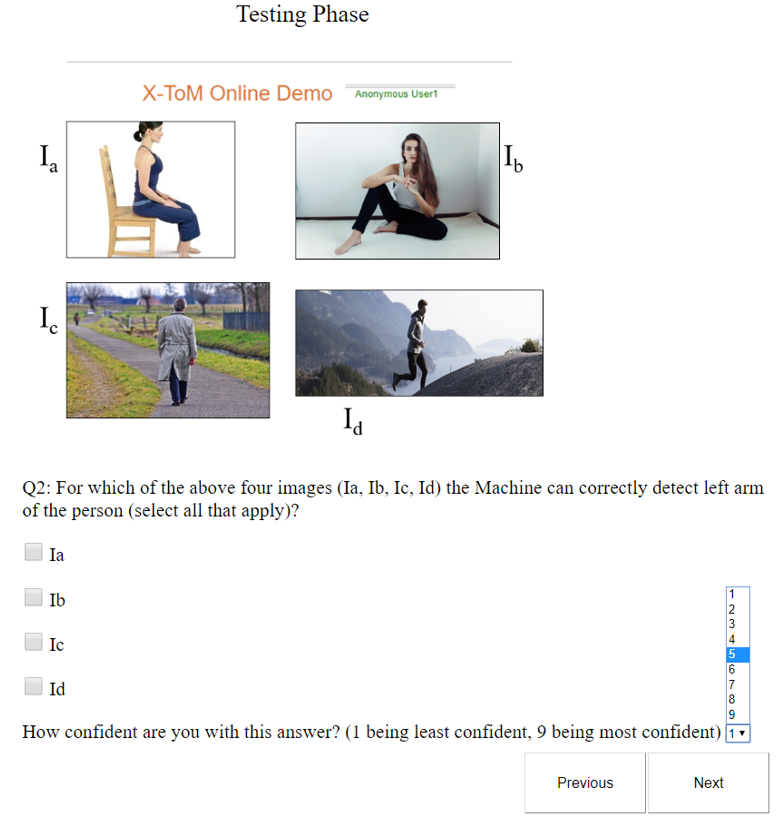}
  \caption{Sample evaluator questions}~\label{fig:sample1}
\end{figure*}

\begin{figure}[!ht]
\centering
  \includegraphics[width=\linewidth]{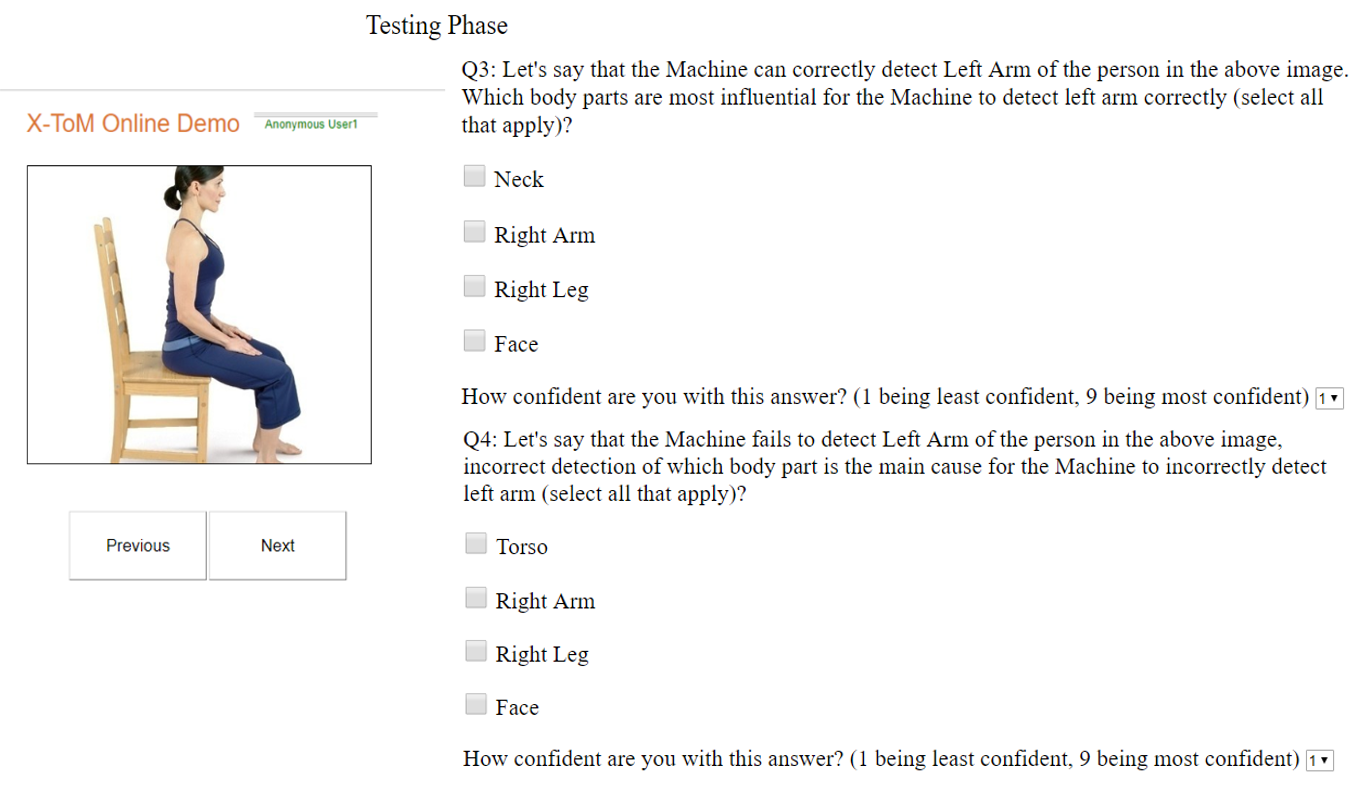}
  \caption{Sample evaluator questions}~\label{fig:sample2}
\end{figure}

\begin{figure*}[!ht]
\centering
  \includegraphics[width=\linewidth]{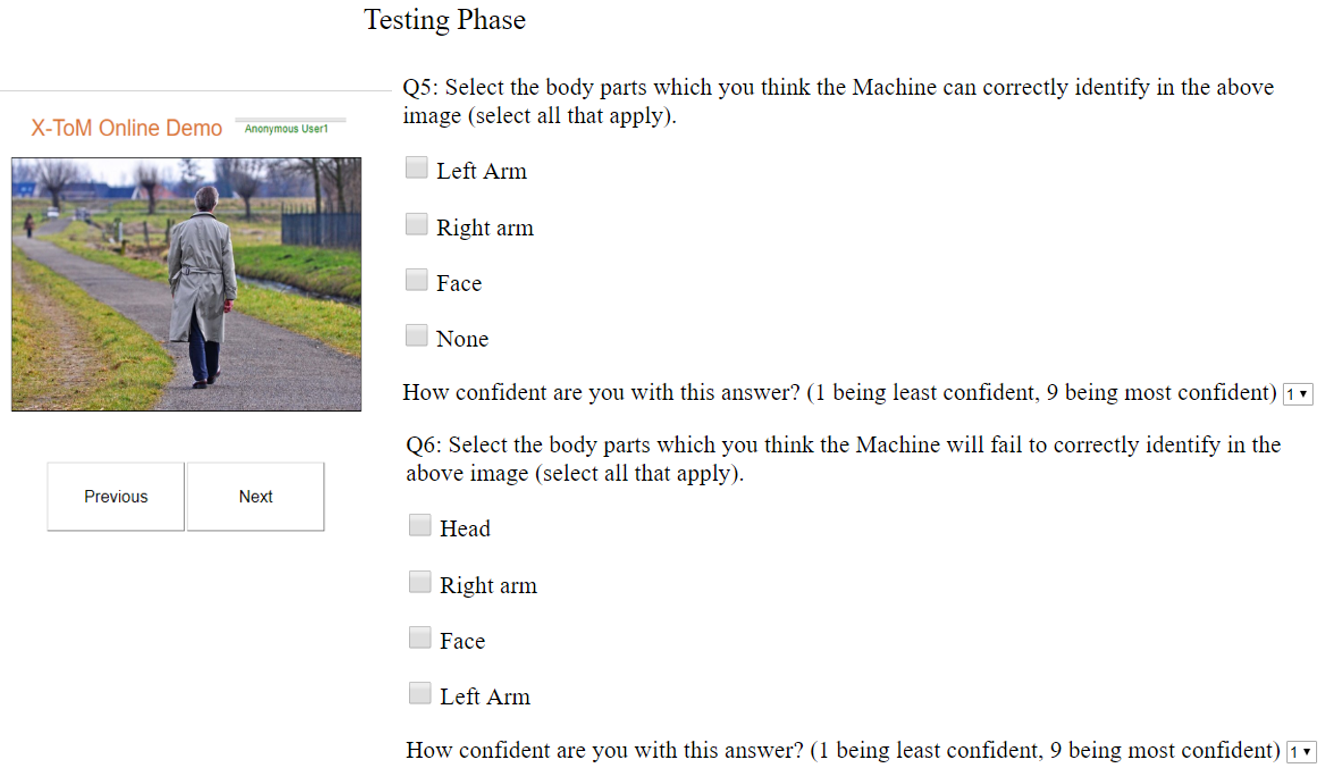}
  \caption{Sample evaluator questions}~\label{fig:sample3}
\end{figure*}

Specifically, there are two main types of evaluator questions about the user’s prediction: (1) whether the Performer would successfully or incorrectly detect objects, parts and other concepts encoded by AOG; and (2) which image parts are most influential for the Performer’s successful or incorrect object detection. For example, the evaluator's questions include
 ``which parts of the image are most important for the machine to recognize that the person is running'', and ``which small part of image contributes most to inferring the surrounding larger part of image''. \Cref{fig:sample1,fig:sample2,fig:sample3} show few sample screenshots (from our web interface) of the exact questions, on the detection of the body part ``Left-Arm", that we pose to the subjects.

\end{document}


\maketitle
\appendix
\section{Appendix}

\subsection{X-ToM Evaluator Interface and Questions}
\begin{figure*}[ht]
\centering
  \includegraphics[width=0.8\linewidth]{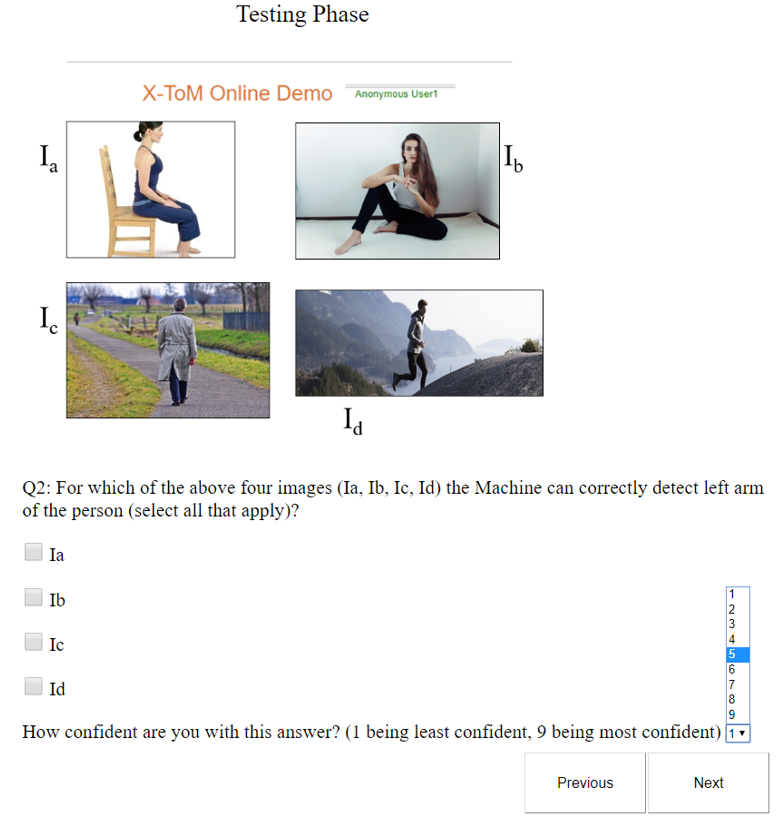}
  \caption{Sample evaluator questions}~\label{fig:sample1}
   \vspace{-15pt}
\end{figure*}

\begin{figure*}[ht]
\centering
  \includegraphics[width=0.95\linewidth]{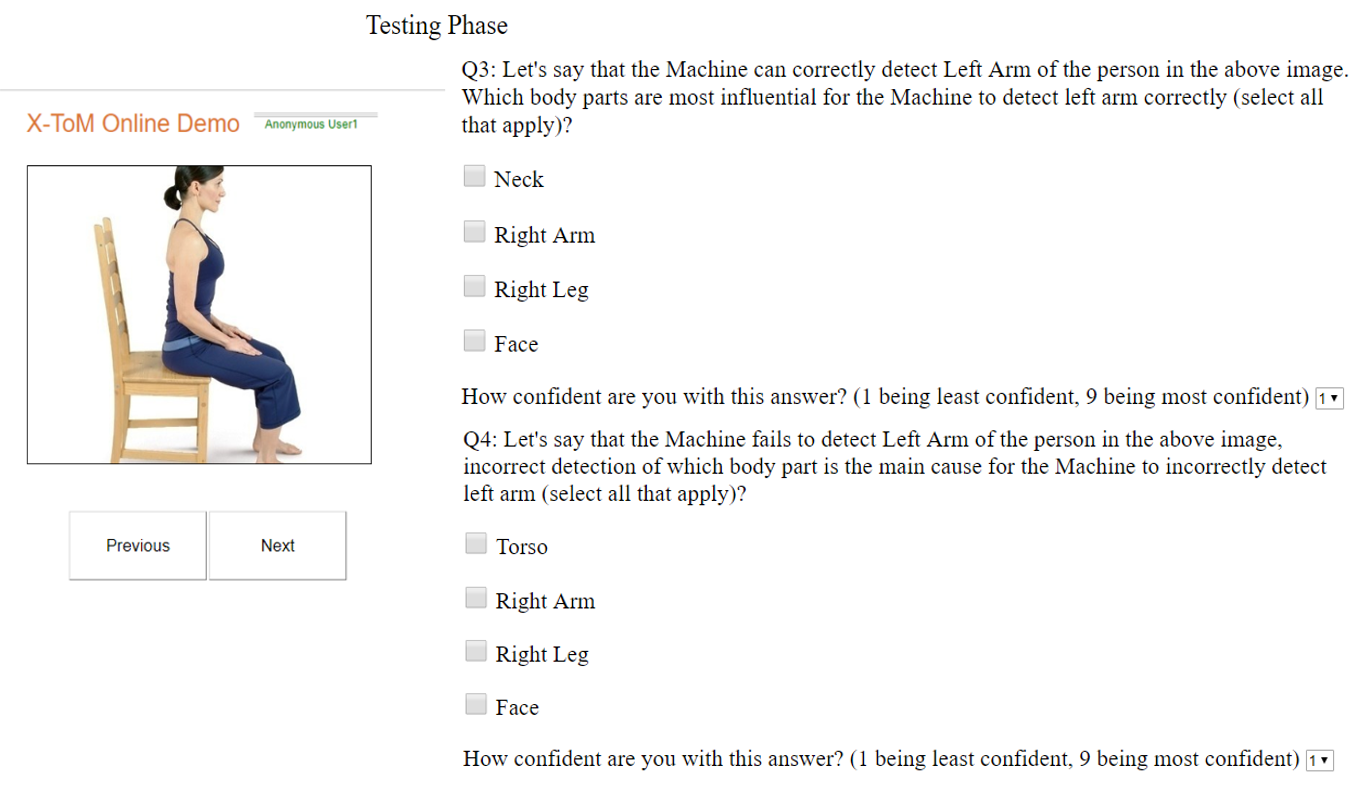}
  \caption{Sample evaluator questions}~\label{fig:sample2}
   \vspace{-15pt}
\end{figure*}

\begin{figure*}[ht]
\centering
  \includegraphics[width=0.95\linewidth]{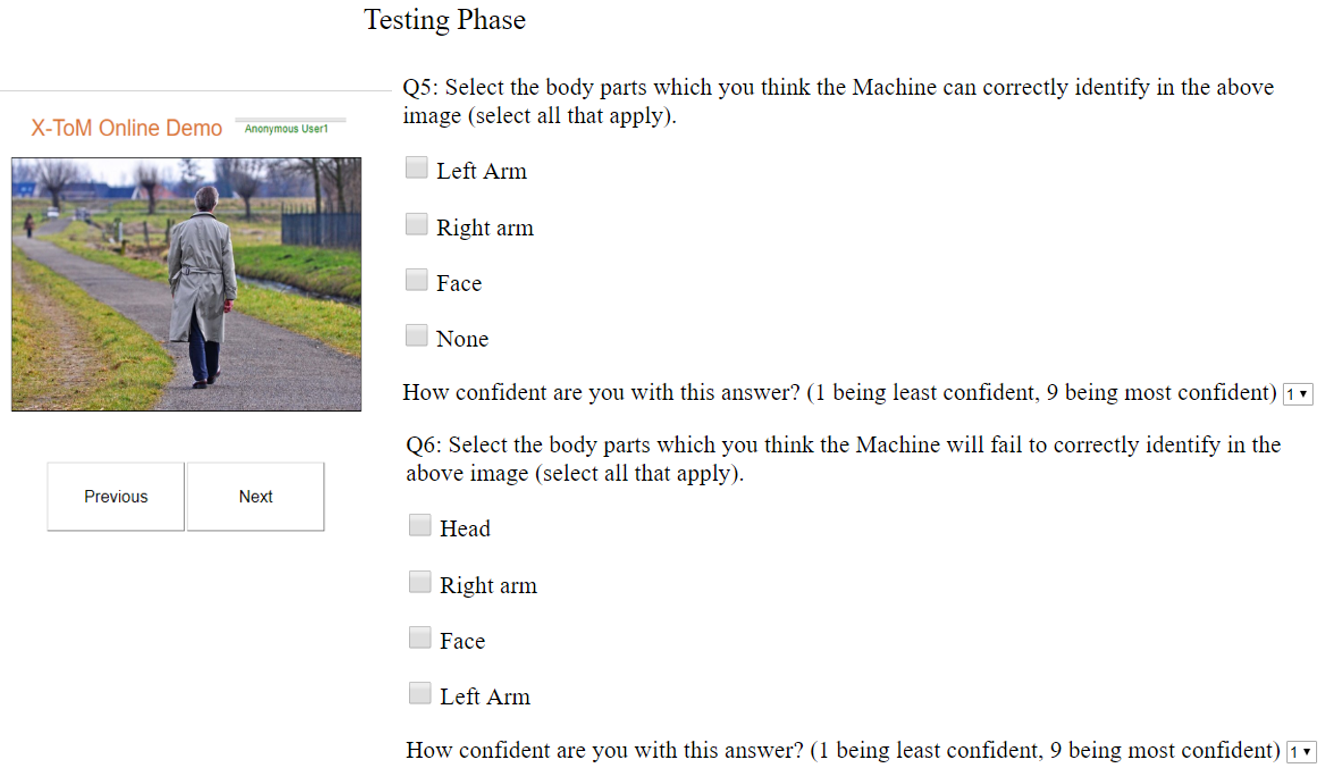}
  \caption{Sample evaluator questions}~\label{fig:sample3}
   \vspace{-15pt}
\end{figure*}

Specifically, there are two main types of evaluator questions about the user’s prediction: (1) whether the Performer would successfully or incorrectly detect objects, parts and other concepts encoded by AOG; and (2) which image parts are most influential for the Performer’s successful or incorrect object detection. For example, the evaluator's questions include
 ``which parts of the image are most important for the machine to recognize that the person is running'', and ``which small part of image contributes most to inferring the surrounding larger part of image''. \Cref{fig:sample1,fig:sample2,fig:sample3} show few sample screenshots (from our web interface) of the exact questions, on the detection of the body part ``Left-Arm", that we pose to the subjects.


\subsection{Evaluation with Psychology Subject Pool}
\begin{figure*}[t]
\centering
  \includegraphics[width=0.95\linewidth]{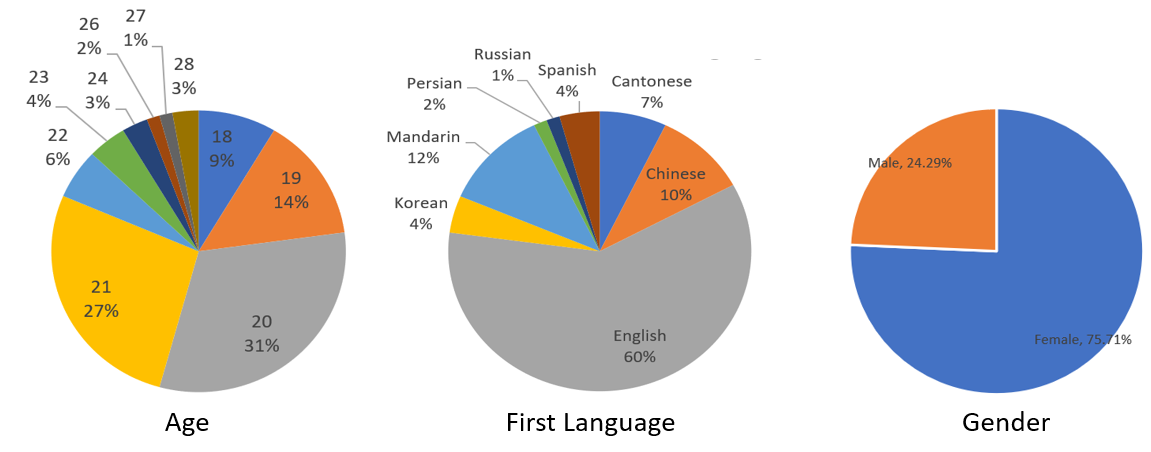}
  \caption{Statistics (based on Age, First Language and Gender) of the 120 human subjects, from Psychology subject pool, participated in our study.}~\label{fig:stats}
\end{figure*}

Figure~\ref{fig:stats} shows the statistics (Age, First Language, Gender) of the 120 human subjects, recruited from our institution's Psychology subject pool.

\subsection{Human Subject Evaluation: Additional Results}

In addition to the metrics Justified Trust and Reliance, we also measure the following metrics for comparing our X-ToM framework with the baselines (QA and Saliency Maps):
\begin{itemize}
\begin{figure}[ht]
\centering
  \includegraphics[width=0.6\linewidth]{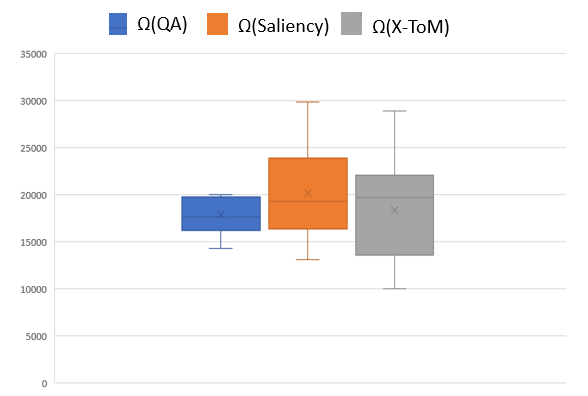}
  \caption{\textbf{Response Times} (in milliseconds per question). Error bars denote standard errors of the means.}~\label{fig:responsetime}
 
\end{figure}
\item \textbf{Response Time}: We record the time taken by the human subject in answering evaluator questions. Figure~\ref{fig:responsetime} shows the average response times (in milliseconds per question) for each of the three groups (X-ToM, QA and Saliency Maps). We expected the participants in X-ToM group to take less time to respond compared to the baselines. However, we find no significant difference in the response times across the three groups.


\begin{figure}[h]
\centering
  \includegraphics[width=0.6\linewidth]{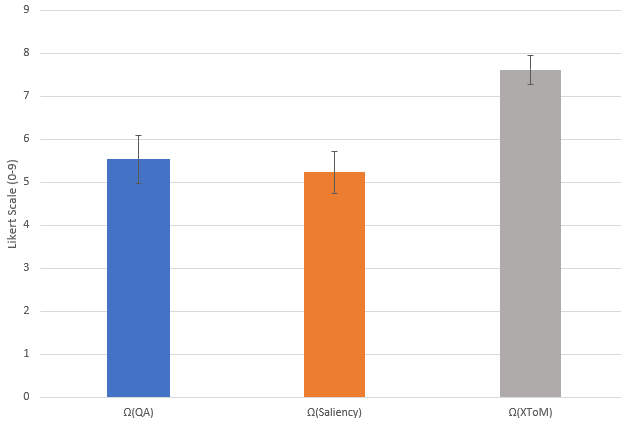}
  \caption{\textbf{Qualitative Reliance}. Error bars denote standard errors of the means.}~\label{fig:subjective}
\end{figure}

\item \textbf{Explanation Satisfaction}:
We measure human subjects' feeling of satisfaction at having achieved an
understanding of the machine in terms of usefulness, sufficiency, appropriated detail, confidence, understandability, accuracy and consistency~\cite{hoffman17explanation,hoffman2018metrics,miller2018explanation,hoffman2010metrics}. We ask them to rate each of these metrics on a Likert scale of 0 to 9. Figure~\ref{fig:satisfaction} shows the average explanation satisfaction rates obtained from each of the three groups. As we can see, subjects in X-ToM experiment group found that explanations were highly useful, sufficient and detailed compared to the baselines ($p < 0.01$). However, we did not find significant differences across the three groups in terms of other satisfaction measures: confidence, understandability, accuracy and consistency.

\item \textbf{Subjective Evaluation of Reliance}: We collect subjective Reliance values (on a Likert scale of 0 to 9) from the subjects in the three groups. The results are shown in Figure~\ref{fig:subjective}.  These results are consistent with our quantitative reliance measures. It may be noted that subjects' qualitative reliance in Saliency Maps is lower compared to the QA baseline.
\\
\\
(\textbf{NOTE}: Please see the figures shown in the next few pages.)
\end{itemize}

\clearpage
\bibliography{naaclhlt2019}
\bibliographystyle{acl_natbib}